# A Service-Oriented Architecture for Assisting the Authoring of Semantic Crowd Maps

Henrique Santos and Vasco Furtado

Universidade de Fortaleza - UNIFOR, Fortaleza CE 60.811-905, Brazil
PPGIA – Programa de Pós-Graduação em Informática Aplicada
`hensantos@gmail.com; vasco@unifor.br`

**Abstract.** Although there are increasingly more initiatives for the generation of semantic knowledge based on user participation, there is still a shortage of platforms for regular users to create applications on which semantic data can be exploited and generated automatically. We propose an architecture, called Semantic Maps (*SeMaps*), for assisting the authoring and hosting of applications in which the maps combine the aggregation of a Geographic Information System and crowd-generated content (called here crowd maps). In these systems, the digital map works as a blackboard for accommodating stories told by people about events they want to share with others typically participating in their social networks. *SeMaps* offers an environment for the creation and maintenance of sites based on crowd maps with the possibility for the user to characterize semantically that which s/he intends to mark on the map. The designer of a crowd map, by informing a linguistic expression that designates what has to be marked on the maps, is guided in a process that aims to associate a concept from a common-sense base to this linguistic expression. Thus, the crowd maps start to have dominion over common-sense inferential relations that define the meaning of the marker, and are able to make inferences about the network of linked data. This makes it possible to generate maps that have the power to perform inferences and access external sources (such as DBpedia) that constitute information that is useful and appropriate to the context of the map. In this paper we describe the architecture of *SeMaps* and how it was applied in a crowd map authoring tool.

## 1 Introduction

Motivated by the huge success of Wikipedia, wiki applications have not been restricted to crowdsourcing via text sharing. On the contrary, there has recently been an explosion of interest in using the Web to create, assemble, and disseminate geographic information provided voluntarily by individuals. Crowd mapping activity, combining the aggregation of a Geographic Information System (maps on the Web) and crowd-generated content, flourishes daily [1, 2]. Sites such as *Wikimapia* (http://www.wikimapia.com), *WikiCrimes* (http://www.wikicrimes.org) [3], Click2fix (http://www.click2fix.co.sa), *Crowdmap* (www.crowdmap.com), and *OpenStreetMap* (http://www.openstreetmap.org) are empowering citizens to create a global patchwork of geographic information, while Google Earth and other virtual globes are encouraging volunteers to develop interesting applications using their own data. In crowd map applications, the digital map works as a blackboard for accommodating stories told by people about events they want to share with others typically participating in their social networks.

On the other hand, although there are increasingly more initiatives for the generation of semantic knowledge based on user participation (e.g.: twine, semantic wiki [4], and ontowiki [5]), there is still a shortage of platforms for the development of applications by non-expert users on which semantic data can be generated automatically and exploited by these applications.

Our work fits innovatively into this context. We propose a service-oriented architecture, called *SeMaps* (from Semantic Maps), for expressing the semantics of what the designer of the map intends to mark on the map (here called markers). We call "semantic characterization" the act of describing the concept(s) that best represent(s) a marker. The characterized concepts are associated with the linked data represented in RDF. In doing so, *SeMaps* makes it possible to generate Semantic Crowd Maps that have the power to perform inferences and/or access external sources that constitute useful and appropriate information to the map context. Wikipedia itself – through its RDF representation in DBpedia [6] – can be one of these external sources that provide additional and contextual information within the map.

In this paper we describe the architecture of *SeMaps*, with an emphasis on the specification of its own ontologies as well as its connection with Linked Open Data (LOD). Then we describe how the *SeMaps* services were coupled to a crowd mapping authoring tool in order to drive the marker's concept elicitation from the commonsense base in order to define the semantic value of what the designer wishes to mark on the map. An example of a crowd map generated from this approach demonstrates how easy the process of attributing semantics and accessing LOD in crowd maps can be.

## 2      Providing Semantics to Crowd Maps

At the core of *SeMaps* are the ontologies that describe the knowledge behind the maps created. Due to the collaboration characteristic inherent in the maps, we sought to reuse concepts of ontology that supported the collaborative model of content creation. We found the support we needed in the Semantically-Interlinked Online Communities (SIOC) ontology [7]. In addition to describing the content created by the participants, it also provides the possibility of adding semantics to such content, an important part of our model. In the *SeMaps* ontology, the marker has information about an event, a person, a business or a particular fact (represented by the concept classes created during the process of marker creation) and on the marker per se (such as date and time, the user who created it, among others). Moreover, in the marker there is a set of features to describe the provenance of the information, such as the type of source and how reliable it is. For representing these latter features, we imported the PML2 (Proof Markup Language) ontology [8]. However, we specialized some of the concepts to cope with the particular features of *SeMaps*, particularly for representing the notion of reputation inspired in [9] and [10]. As shown in Figure 1, the main classes of *SeMaps* are:

- Marker – the class responsible for describing a user's report about something of the domain of the map at issue. The property *sioc:has_creator* links to the user account who reports and the property *sioc:topic* links to the concept being reported.
- WikiUser/WikiUser Account – WikiUser describes a user registered on a particular crowd map. Those users make up a social network through the concepts of person and friends imported from the FOAF (friend of a friend) ontology [11]. WikiUser Account describes the account the user utilizes to interact with the map.
- Concept classes – describes the thing being reported on the crowd map. Instances of this class also link to an *InferenceNet* concept via the *rdf:type* property. As *SeMaps* assumes that a marker must be a person, an organization, an event, a complaint, an artistic production (movies, photos, books, etc.), a building, or a commercial establishment, every created class is a subclass of these classes (we dub them the *SeMaps* top classes).

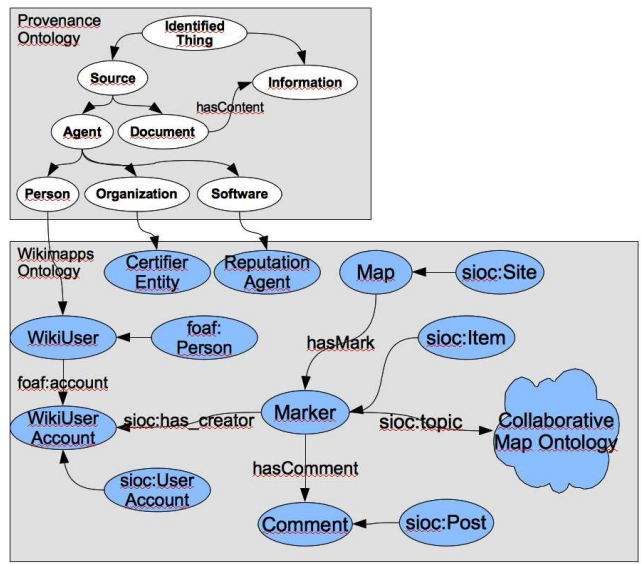

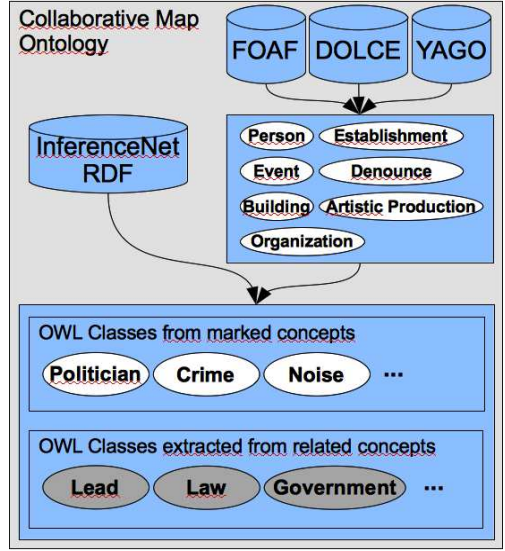

**Fig. 1.** *SeMaps* Ontologies

*SeMaps* allows the generated maps to have completely heterogeneous domains. Therefore, there is a need to describe types of information that exist only for a specific map. A particular map can refer to crimes, noise pollution, humanitarian assistance, etc. Each map contains specific information that must be described. To enable this specialized description, OWL classes are created on demand, i.e., *SeMaps* representation evolves according to the map created. These classes are extracted directly from the markers created on the various crowd maps and are images of the *sioc:topic* property. As previously mentioned, the instances of these classes are linked to a concept through the *rdf:type* property.

Besides using the aforementioned ontologies, *SeMaps* services rely on a bilingual knowledge base, which expresses inferentialist semantic and common-sense knowledge – *InferenceNet* [12]. *InferenceNet* expresses semantic content through a network that connects a concept to many others through dozens of common-sense semantic relations and that are inferential in nature (preconditions and postconditions of the use of concepts). Formally, this base is represented in a directed graph $G_c(C,Rc)$. Each inferential relationship $rc_j \in Rc$ is represented by a tuple (*relationName*, $c_i$, $c_k$, *type*), where *relationName* is the name of an *InferenceNet* semantic relation (Capableof, PropertyOf, EffectOf, etc.), $c_i$ and $c_k \in C$ are concepts of a natural language, and type = "Pre" or "Pos", indicating a precondition or a postcondition for using the concept $c_i$.

*InferenceNet* is linked to YAGO and DBpedia [13], which allows rich inferences to be made, since the base expresses common-sense knowledge and the inferential import of the concepts in the reasoning.

## 3 The *SeMaps* Architecture

*SeMaps* is a set of web services (WS) specifications, also including built-in functionalities, capable of enabling semantics on crowd maps. The main features are: description of crowd maps in an RDF/OWL ontology, web services for semantically characterizing a maker, and web service for LOD contextualized resources retrieval from data already on the map. Figure 2 shows the *SeMaps* architecture and its relation with a crowd map.

### 3.1 Semantic Characterization Module

The module for semantic characterization comprises the services by which the user-designer can define the semantic value s/he wishes to mark on the map (the marker) based on the common-sense base. In short, the user-designer, by informing a linguistic expression that designates its marker, is guided in a process that aims to associate a concept from the common-sense base to the linguistic expression that names the marker. Thus, the map starts to have dominion over common-sense inferential relations that define the meaning of the marker and are able to make inferences about the network of linked data. *SeMaps* provides services to associate a concept to a marker and to create an ontology for the crowd map.

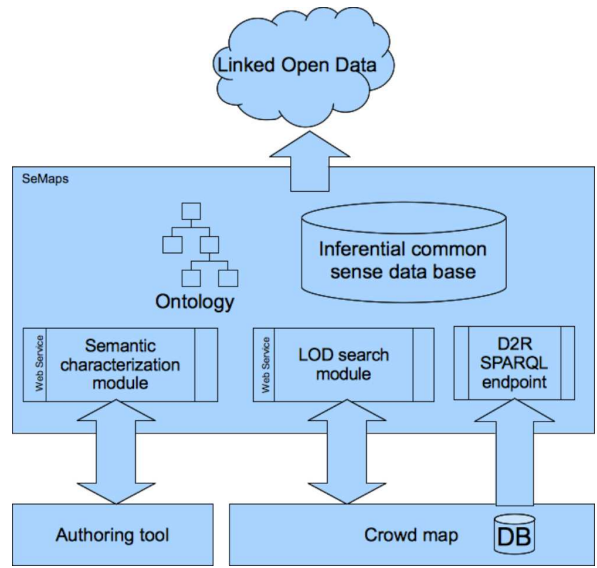

**Fig. 2.** *SeMaps* Architecture

The association of a concept to a marker is an iterative process that occurs while there are markers to be created by the designer. Figure 3 describes the interface of the WS (*SemanticCharacterization*) with its methods and I/O parameters. The *semanticCharacterize* method receives as a parameter the string that describes a marker and returns a list of URIs that identify resources found in the common-sense base represented in RDF.

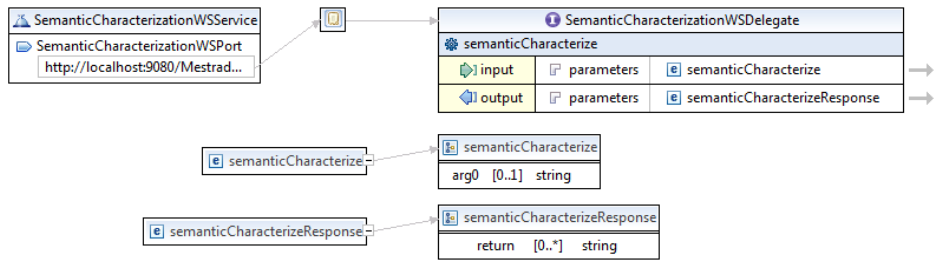

**Fig. 3.** Semantic characterization web service interface

For each marker with an associated concept, *SeMaps* uses another service to create the classes of the ontology for this map. The instances of the classes are linked to DBpedia/YAGO. Figure 4 describes the interface of the Concepts WS. The input parameter of the *createConcept* method is the *string* that describes the marker and returns a unique identifier for the concept. This number is the main reference of the concept in *SeMaps*.

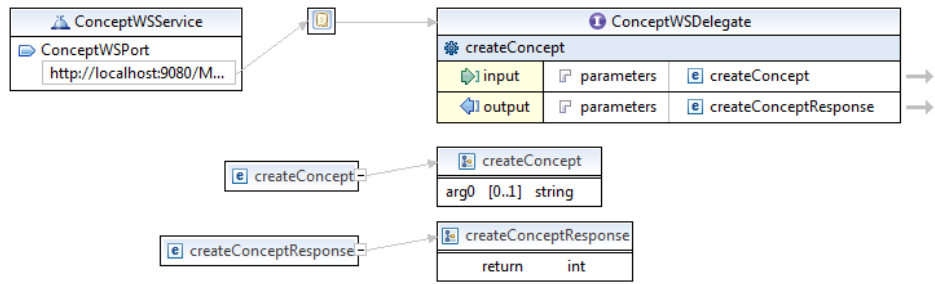

**Fig. 4.** Concept creation module web service interface

### 3.2 LOD Search Module

The search module of *SeMaps* provides a service that obtains resources (articles, news, etc.) from LOD. This search is based on the semantic characterization done in the module described in 3.1. The *searchLODForConcept* method of the LOD WS receives a concept representing a marker as input and returns a list of URIs describing resources related to the concept. Usually the algorithms that implement this search also use *InferenceNet*. An example of this was described in [14]. The URIs returned by the method can be used in different ways, for instance, used as a widget or plotted directly on the map. Due to space constraints, the web service description in WSDL was omitted.

### 3.3 D2R Endpoint

*SeMaps* assumes the use of a service of mapping relational data to RDF. This is necessary not only to avoid data replication, since typically the authoring tools have their data represented in RDBs, but also because the crowd map data will be available to be accessed by a SPARQL endpoint. We used the D2R server [15], which is an open and free system for publishing relational data on the Web. It enables RDF and HTML browsers to navigate the content of non-RDF databases, and allows applications to query a database using the SPARQL query language over the SPARQL protocol.

## 4 Integrating *SeMaps* into WikiMapps

The *SeMaps* architecture allows several different implementations for each service. For example, different algorithms for semantic annotation can be used in the process of semantic characterization. The only requirement that must be satisfied is to follow the interface specified previously.

In order to put *SeMaps* to work, we integrated the *SeMaps* services for semantic characterization of a marker into *Wikimapps* (www.wikimapps.com), an authoring tool for creating and hosting crowd maps. The semantic characterization process and the search module were instantiated with algorithms described in [14].

We also used a D2R Server to maintain the map ontology. Thus, for each marker created in the authoring tool, *SeMaps* creates an entry in this table and a mapping in D2RQ helps to automatically update the ontology.

To exemplify how the knowledge represented on a crowd map generated from *Wikimapps* works, let's assume that a designer wants to create a map of politicians of a certain region. In order to associate semantics to the markers, *Wikimapps* was prepared to call the services for semantic characterization and concept creation of *SeMaps*. Thus, when the designer of a crowd map in *Wikimapps* defines a marker called "politician", it is automatically associated with *InferenceNet*'s "politician" concept that, in turn, is linked to instances of YAGO and DBpedia. The way each service of this process was implemented is described in [14].

The crowd map generated by *Wikimapps* was also prepared to use a *SeMaps* search WS. This service was called inside a widget that brings LOD data from the New York Times and DBpedia, with additional information related to subjects to which the markers of the map refer. The widget calls the WS whenever there is a change in the map's viewport.

Figure 5 presents the results of a query made by the crowd map, whereby news from the New York Times on corruption, scandals and laws has been found. Note that this information refers to the region the map selected (in this case the state of Illinois).

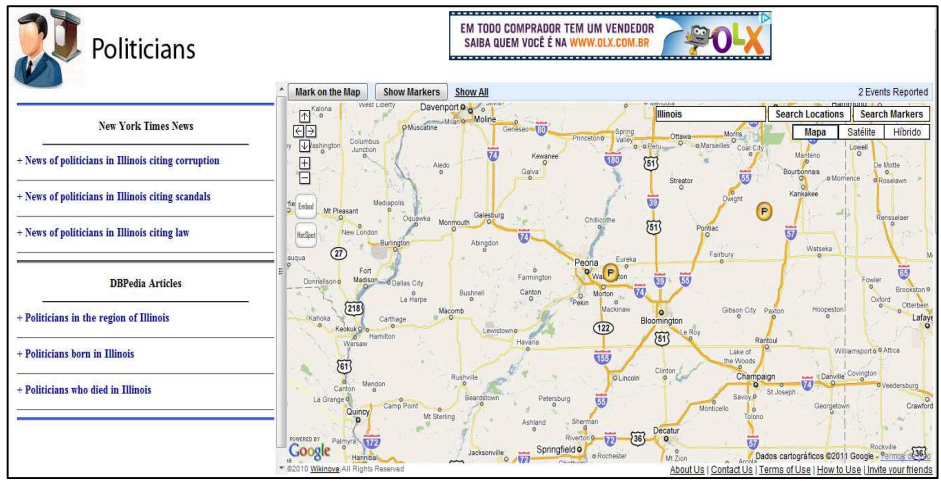

**Fig. 5.** Screen shot of the "Politicians" Crowd Map with News and Articles

Figure 5 also shows links representing the relations of common sense that have been identified. If the user decides to navigate the map, other queries are performed, this time filtering information regarding relationships as news stories on laws proposed by politicians, corruption, or any scandals. In the example shown in Figure 5, the possible issues were created by the inferential content of the concept "politician", expressed in *InferenceNet*: "a politician is able to propose laws" and "a politician has the capacity to corrupt." This common-sense knowledge is expressed in *InferenceNet* through the relationships (CapableOf, "politician", "to propose laws", "Pre") and (ProprietyOf, "politician", "corruption", "Pre"), respectively. Moreover, the concepts "political", "law" and "corruption" are associated – via LOD – with YAGO resources and DBpedia. Also in Figure 5, we see another characteristic obtained by the fact of knowing the semantics of the concept: DBpedia articles are related to the location indicated on the map. Further information regarding the place of birth, death, work and other references that are based on *GeoNames* allow *WikiMapps* to automatically sort the items for these characteristics.

## 5 Related Work

The production of linked open data is a growing trend on the web. Thus, the number of tools for generating semantic content on the Web is increasing. Highlighted among these tools are those that export pre-existing data without semantics, such as Drupal [16].

*LinkedGeoData* [17] is a project aimed at processing and representation of RDF data that was created collaboratively in *OpenStreetMap* (OSM – www.openstreetmap.org). In addition, the project aims to allow binding of OSM with other bases of the LOD. *LinkedGeoData* developed an ontology partly derived from the relational schema coming from *OpenStreetMap* and WGS84 (World Geodetic

System). To map the collected data in instances of the OSM ontology, *LinkedGeoData* makes use of Triplify, a tool able to accomplish mapping from relational databases to RDF. Despite the similarities with *SeMaps*, *LinkedGeoData* is focused on the geographic elements that compose the map instead of modeling denounces, events, persons and other entities that are associated to places. This is not a disadvantage of *LinkedGeoData*, but shows how it is focused on collaborative cartography and it is not able to cover what *SeMaps* covers, like events or complaints, for instance.

Another group of tools that produce semantic content is that of semantic wikis. The semantic wikis, such as *Semantic MediaWiki* [4], allow the users themselves to describe – through notes – the content edited by them. Although they allow semantic data to be generated collaboratively, semantic wikis tend to make the process of creation somewhat difficult for users, since they need to use special tags in order for the data to be linked semantically. This difficulty can result in data linked erroneously or without sufficient links. Some projects such as Wolfram Alpha (www.wolframalpha.com) aim to provide a "computational knowledge engine" and allow the generation of knowledge widgets that are very similar to the services that can be produced using the services provided by *SeMaps*. However, a proprietary knowledge base is used, exploring alternative representational ways that could eventually support or complement our approach.

Hermes [18] is a system capable of creating a knowledge base and from there, filtering news from RSS feeds in accordance with the desired concepts. It compiles an ontology with linguistic expressions that appear in the selected news items. By creating a knowledge base on top of news, the proposed framework does not use a knowledge base of common sense as we do in *SeMaps*.

## 6  Conclusion

In this paper we propose the semantic crowd map concept, in which the process of participation and interaction always occurs in relation to a given space indicated on a digital map and the semantic of the marker is provided by services of *SeMaps*. We describe a platform for creating semantic crowd maps, which provides an environment for the creation and maintenance of sites based on crowd maps. We seek to supply the lack of platforms for the development of applications, by non-specialist users, where semantic data are automatically generated and exploited by these applications.

*SeMaps* allows a user who wishes to create a crowd map to do so with the description of the concept of that which s/he wishes to mark on the map. Hence, the maps produced by *SeMaps* have the power to make inferences and access external sources that constitute useful information, appropriate to the context of the map. This semantic characterization of the marker used a common-sense base in Portuguese and English. The integration of *SeMaps* into *Wikimapps* showed that crowd maps generated by *WikiMapps* are more informative, since before *SeMaps* they only were able to show what people marked on the map and nothing more. With *SeMaps*, they are able to access LOD and show relevant and contextualized information.